\pdfoutput=1

\documentclass[letterpaper, 10pt, conference]{ieeeconf}

\IEEEoverridecommandlockouts                       %
\usepackage[utf8]{inputenc}
\usepackage[T1]{fontenc}
\usepackage{graphicx}
\usepackage{booktabs}
\usepackage{multirow}
\usepackage{amsmath,amssymb,amsfonts}
\usepackage{mathtools}
\usepackage{makecell}
\usepackage{xcolor}
\usepackage{microtype}
\usepackage[hidelinks]{hyperref}
\usepackage[capitalize,noabbrev]{cleveref}
\hypersetup{pdftitle={Perceives but Doesn't Brake: A Perception-Action Binding Gap in Reasoning Driving VLAs},
  pdfauthor={Matteo Cercola, Katie Z Luo, Simone Formentin, Marco Pavone}}

\newcommand{\alpamayo}{Alpamayo-1.5}
\newcommand{\vavam}{VaVAM}
\newcommand{\lmdrive}{LMDrive}
\newcommand{\senna}{Senna}

\newcommand{\ci}[2]{(#1,\,#2)}                     %
\newcommand{\dsafe}{d_{\mathrm{safe}}}
\newcommand{\vtarget}{v_{\mathrm{target}}}
\newcommand{\sfp}{\operatorname{sp}}                %

\newcommand{\CachedNHaz}{144}

\newcommand{\CachedRangeMAE}{2.20}

\newcommand{\NewNHaz}{91}

\newcommand{\NewRangeMAE}{3.28}

\newcommand{\CachedRangeMAETele}{1.23}
\newcommand{\CachedRangeMAEAttn}{1.99}
\newcommand{\NewRangeMAETele}{2.43}
\newcommand{\AttnRangeMAEBare}{2.21}

\newcommand{\NewIntrUnguidedPct}{65.0\%}
\newcommand{\NewIntrOraclePct}{6.0\%}
\newcommand{\AttnIntrPct}{11.7\%}
\newcommand{\NativeSetN}{273}
\newcommand{\NativeSetClips}{185}
\newcommand{\NativeReachSpeedRho}{{+}0.645}
\newcommand{\NativeReachSpeedRhoCI}{\ci{{+}0.552}{{+}0.721}}
\newcommand{\NativeReachRangeRho}{{+}0.009}
\newcommand{\NativeReachRangeRhoCI}{\ci{$-$0.110}{{+}0.125}}
\newcommand{\NativeGuidedReachRangeRho}{{+}0.678}
\newcommand{\NativeGuidedReachRangeRhoCI}{\ci{{+}0.599}{{+}0.744}}
\newcommand{\NativeGuidedReachSpeedRho}{{+}0.177}
\newcommand{\NativeGuidedReachSpeedRhoCI}{\ci{{+}0.025}{{+}0.337}}
\newcommand{\NativeSlopeUnguided}{$-$0.261}
\newcommand{\NativeSlopeUnguidedCI}{\ci{$-$0.905}{{+}0.283}}
\newcommand{\NativeSlopeGuided}{{+}0.474}
\newcommand{\NativeSlopeGuidedCI}{\ci{{+}0.142}{{+}0.921}}
\newcommand{\NativeIntrusionMeanK}{2.50\%}

\let\oldthebibliography\thebibliography
\renewcommand{\thebibliography}[1]{\oldthebibliography{#1}\setlength{\itemsep}{0pt plus 0.3pt}}
\renewcommand{\baselinestretch}{0.973}

\title{\LARGE \bf Perceives but Doesn't Brake: A Perception--Action Binding Gap\\ in Reasoning Driving VLAs}

\author{Matteo Cercola$^{1}$, Katie Z Luo$^{2}$, Simone Formentin$^{1}$ and Marco Pavone$^{2,3}$%
\thanks{This work has been submitted to the IEEE for possible publication. Copyright may be transferred without notice, after which this version may no longer be accessible.}%
\thanks{$^{1}$Politecnico di Milano. $^{2}$Stanford University. $^{3}$NVIDIA.}%
\thanks{Project page: \url{https://matcercola18.github.io/pbdb/}}%
}

\begin{document}

\maketitle
\thispagestyle{empty}
\pagestyle{empty}

\begin{figure*}[t]
  \centering
  \includegraphics[width=0.88\textwidth]{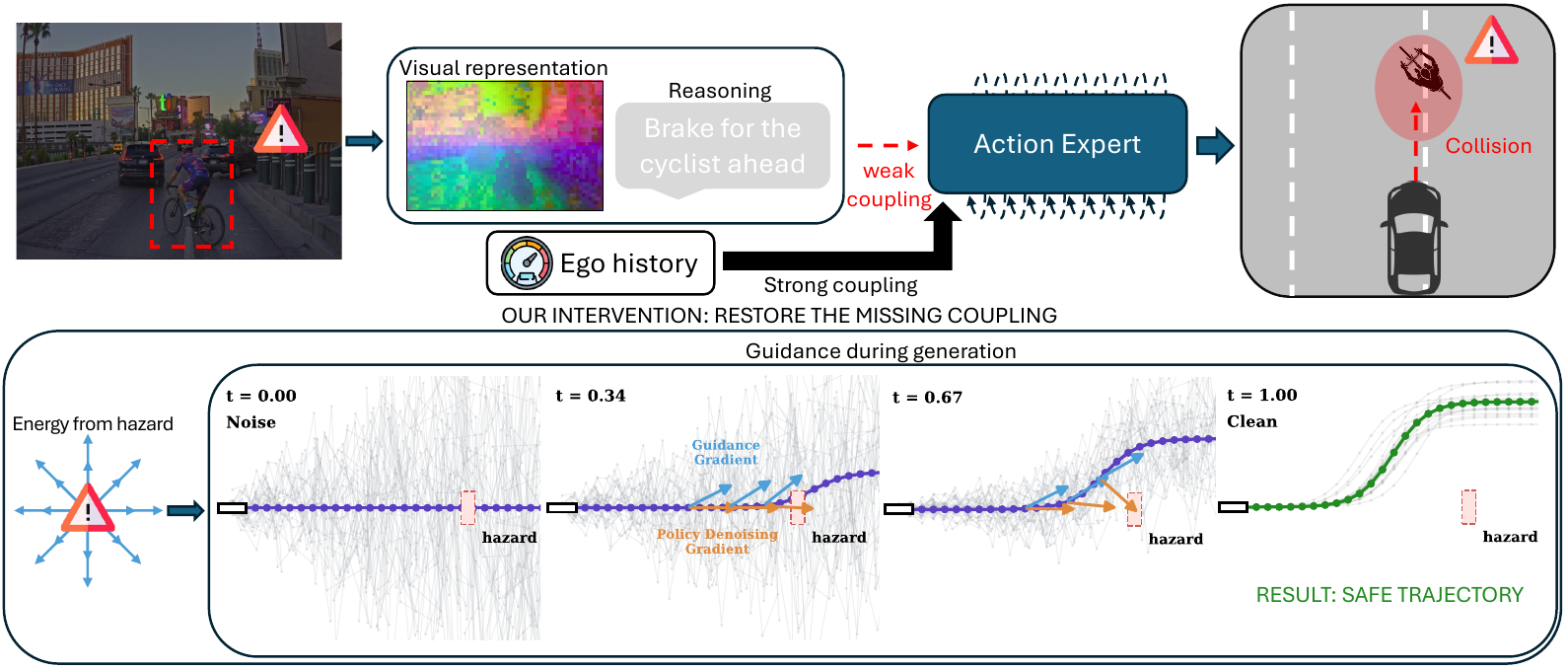}
  \caption{\textbf{Where perception--action binding fails, and where we intervene.}
  \emph{Binding gap:} visual hazard evidence and the model's reasoning trace
  have substantially weaker authority over the longitudinal plan than ego
  history. \emph{Intervention:} a hazard-conditioned energy gradient is added
  during flow integration, steering the frozen trajectory generator toward a
  safe plan without modifying model weights.}
  \label{fig:hero}
\end{figure*}

\begin{abstract}
Reasoning-augmented vision-language-action (VLA) driving models can perceive and reason about safety-critical hazards yet fail to translate that evidence into safe actions. We characterize this failure as a \emph{perception--action binding gap}: hazard information is available inside the model but exerts insufficient control over trajectory generation. Using matched simulated counterfactuals and logged driving replay, we show that in-path hazards are actually encoded in the visual representation, but longitudinal plans are dominated by ego-speed history and respond only weakly to hazard proximity or the model's own reasoning trace. Further analysis shows that the trajectory generator is capable of braking but sometimes fails to do so due to the perception-action binding gap. Without retraining or modifying any weights, we inject a hazard-conditioned energy into the frozen flow-matching sampler of \alpamayo{}, restoring the missing coupling at generation time. With exact obstacle geometry, guidance cuts footprint intrusion from 58.1\% to 14.5\% and recovers safe braking whenever sufficient stopping distance remains, eliminating collisions entirely in our braking-relevant closed-loop battery. It also reverses the diagnosed dependence of the plan, from tracking ego speed to responding to obstacle range. These results separate hazard perception from hazard control in driving VLAs, show that unsafe behavior can arise even when the model has both the evidence and the ability to act on it, and propose a simple fix for bridging this binding gap.
\end{abstract}

\section{Introduction}
\label{sec:intro}

Reasoning-augmented vision--language--action (VLA) models are increasingly
designed to do more than map observations directly to actions: they perceive
the scene, produce an explicit reasoning trace, and then generate a driving
trajectory. This architecture suggests a natural safety premise: if the model
correctly identifies a hazard and reasons that the vehicle should respond, that
evidence should influence the resulting action. 
However, on simulated scenes containing a stationary obstacle in the ego
lane, a state-of-the-art
reasoning VLA~\cite{alpamayo2025r1} generates trajectories that enter the
obstacle footprint on 58.1\% of samples, and its selected actions collide
with a reconstructed version of that obstacle in most closed-loop rollouts, even though a comfortable stop was still possible. Crucially, the failure does
not arise because the hazard is absent from the model. The obstacle is
recoverable from the visual representation and is often explicitly
identified in the model's own reasoning, yet neither source of evidence
exerts sufficient influence on the generated longitudinal plan. Instead,
the plan remains strongly governed by the supplied ego-speed history. We
call this dissociation between accessible hazard evidence and its influence
on action a \emph{perception--action binding gap} (\Cref{fig:hero}); here
\emph{perceives} means recoverable from the model's representation
(formalized as \emph{availability}, \Cref{sec:diagnosis}), not awareness.

This distinction matters because an unsafe action can have at least three
different causes. First, the relevant hazard may not be represented at all
(\emph{availability}). Second, the trajectory generator may be incapable of
producing an appropriate response even when the hazard is known
(\emph{action-side sufficiency}). Third, the evidence may be available and a
safe action may be expressible, but the evidence may have insufficient
\emph{action authority}: changing the hazard does not sufficiently change the
generated plan. These cases imply different remedies, yet standard driving
metrics typically conflate them.

We disentangle them with matched simulated interventions that vary
visual hazard content, obstacle range, ego-speed history, and textual reasoning
while holding competing inputs fixed. Hazard information is available
throughout, yet longitudinal behavior is far more sensitive to ego history than
to the hazard itself, under closed-loop control in AlpaSim/NuRec \cite{alpasim_2025} as well as
open-loop scoring.

Next, we ask whether the missing coupling can be restored without retraining the
policy. We inject a differentiable hazard energy into the frozen
flow-matching sampler of \alpamayo{}, steering trajectory generation during
flow integration while leaving all model parameters unchanged. Supplied exact
obstacle geometry, this reduces footprint intrusion from
58.1\% to 14.5\% and changes the governing dependence of the generated plan:
trajectory reach becomes primarily responsive to obstacle range rather than ego
speed. Controlled CARLA sweeps further show that the remaining failures are
those in which insufficient stopping distance remains.

Finally, we ask how far this intervention can be driven by signals
already present in the frozen model, rather than by privileged geometry.
Lightweight readouts of the visual cache estimate both when an in-path
hazard is present and where it lies. The key is to preserve the spatial
structure needed for that estimate: on sites held out in both geometry
and appearance, a spatially aware readout reduces footprint intrusion
from \NewIntrUnguidedPct{} to \AttnIntrPct{}, approaching the
\NewIntrOraclePct{} obtained with exact geometry, with no measured
reach loss on matched clear scenes. Thus, much of the safety constraint
needed to restore perception-action binding is already recoverable from the model's own
representation. 
Autonomous binding is therefore feasible within
the evaluated visual domain, while robust semantic triggering under
domain shift remains open.

Our contributions are threefold. \textbf{First, our primary contribution is to identify and causally diagnose a perception--action binding gap in a reasoning driving VLA.}
Across logged driving, matched counterfactuals, and closed-loop replay, we
separate hazard availability from action authority and show that
longitudinal planning is dominated by ego-speed history; to our knowledge,
this is the first causal, rather than correlational, separation of hazard
availability from action authority in a reasoning driving VLA. \textbf{Second,
we show that the missing coupling can be enforced at inference time without
retraining.} A differentiable hazard-conditioned energy steers the frozen
flow-matching sampler toward safe braking and, when supplied an accurate
constraint, shifts the plan's dependence from ego speed toward obstacle
range, demonstrating action-side sufficiency. \textbf{Third, we show why standard logged evaluation can miss such a
failure.} Because a binding gap is a failure of hazard-conditioned action response rather
than average prediction quality, standard logged evaluation can miss it:
safety-critical states in which a moving ego must brake for a stationary
in-path obstacle are rare because human drivers typically avoid them.

\section{Related Work}
\label{sec:related}

\subsection{Ego-State Shortcuts and Driving Evaluation}
Prior work shows that open-loop driving benchmarks can be solved surprisingly well from ego dynamics alone~\cite{zhai2023admlp}, and that planners remain nearly unchanged under severe visual corruption while responding strongly to ego-velocity perturbations~\cite{li2024ego}. These findings establish an ego-state shortcut but do not distinguish whether visual evidence is absent from the representation or present yet uncoupled from action.

\subsection{Reasoning and Video--Action Driving Models}
Recent driving policies increasingly couple visual observations, language or
reasoning, and action generation. \alpamayo{} combines chain-of-causation
reasoning with a flow-matching trajectory expert~\cite{alpamayo2025r1};
\senna{} predicts language-level driving decisions~\cite{senna2024};
\lmdrive{} combines language, perception, navigation, and explicit ego
state~\cite{shao2024lmdrive}; and \vavam{} is a language-free video--action
policy without explicit ego-state input~\cite{bartoccioni2025vavim}.

\subsection{Reasoning--Action Faithfulness}
Prior work shows that stated reasoning and executed behavior can disagree,
including within the same model family~\cite{cotfaithful2026av}, and that
embodied reasoning faithfulness can be improved with learned
critics~\cite{pinocchio2026}. These studies show language and action can
diverge, but do not separate hazard availability from whether it exerts
causal authority over the trajectory.

\subsection{Inference-Time Guidance for Generative Policies}
Gradient guidance can steer pretrained diffusion and flow samplers during
generation without retraining~\cite{dhariwal2021diffusion,lipman2023flow,guidedflows2023};
runtime reasoning--action alignment has also used sample-and-rerank
schemes~\cite{dowhatyousay2025}. We use guidance as a diagnostic intervention rather than proposing a new
control mechanism: by granting a safety constraint direct authority during sampling, we test whether the frozen generator can produce a response that its native conditioning fails to elicit.

\section{Background}
\label{sec:setup}

\subsection{Model and Evaluation Instruments}

We evaluate \alpamayo{}~\cite{alpamayo2025r1} frozen and zero-shot, preserving its native input and ego history conditioning. The flow-matching trajectory expert generates a 6.4\,s trajectory; unless otherwise noted, we sample $K{=}6$ trajectories per state.
All instruments study the same critical state: an approaching ego vehicle
facing a stationary in-path obstacle.

\paragraph{Logged scenes}
From held-out US test clips of the NVIDIA PhysicalAI-AV
set~\cite{nvidia2025physicalai}, we select states where ego speed is at least
4\,m/s and a stationary object lies 8--45\,m ahead in the ego corridor,
using only contemporaneous labels and never the driver's future trajectory,
yielding $n{=}\NativeSetN{}$ states from \NativeSetClips{} clips. These states form a \emph{nominal diagnostic set}, not a hazard benchmark: the objects are predominantly vehicles parked at the curb, no braking is required of the ego, and at the model's native input its plans enter the footprint on only \NativeIntrusionMeanK{} of samples. We use these scenes only to measure which input variable the plan's forward reach tracks.

\paragraph{Designed hazards in controlled simulation}
Using CARLA~\cite{dosovitskiy2017carla} through
Fail2Drive~\cite{fail2drive2026}, we construct \emph{matched-despawn pairs}:
identical sites, rendering, ego motion, history, and native four-camera
temporal input, differing only in whether the stationary obstacle is
present, isolating the causal effect of hazard content. Hazards span six
object kinds at 8--30\,m. We report two populations: \emph{seen sites}
($n{=}\CachedNHaz{}$ hazard and 24 empty scenes over six sites, evaluated
leave-one-site-out) and \emph{new sites} ($n{=}\NewNHaz{}$ and 12 over six
disjoint sites held out in both geometry and appearance). Unless stated
otherwise, 95\% confidence intervals (CIs) are site-clustered bootstraps
($B{=}10^4$).

\paragraph{Closed-loop replay}
In AlpaSim/NuRec, we test repeated replanning with the native four-camera
\alpamayo{} controller, using two stimuli. In the \emph{static} one we
re-pose a reconstructed vehicle 40\,m ahead in the ego lane and freeze it,
paired with its unedited reconstruction as control; 24 of 27 candidate
scenes completed both rollouts. In the \emph{dynamic} one, we instead use a lead vehicle already present in the simulation. For each of 38 scenes in which this vehicle remains ahead of the ego in the same lane, we compare two matched rollouts: in the \emph{Drives on} control, it continues along its logged motion, while in the \emph{Stops} condition, we stop it at a pose it actually occupies in the logged scene. Thus, unlike the static stimulus, no vehicle is inserted or repositioned; only the lead vehicle's subsequent motion is changed.

For the static stimulus, placing the obstacle 40\,m ahead only initializes
the encounter: the policy then replans repeatedly as the ego approaches it.
Crucially, the failure occurs while braking is still physically feasible.
At the relevant replanning states, the obstacle is 8--24\,m ahead and stopping would require only 0.35--1.4\,m/s$^2$ of deceleration. 

\subsection{Metrics}
\label{sec:metrics}

Our primary open-loop metric is \emph{footprint intrusion}: the fraction of
sampled trajectories whose swept ego footprint intersects the obstacle's. We
also report \emph{forward reach}, the
distance traveled before rest (or over the full horizon), and \emph{minimum
clearance}, the smallest separation between swept ego and obstacle
footprints, signed negative when they overlap. For close-range CARLA
experiments, \emph{near-stop} denotes forward reach below 8\,m.

For the static closed-loop stimulus we report the \emph{front-collision
rate}: rollouts whose ego forward footprint edge overlaps another agent's or
the obstacle's footprint, distinguishing forward from rear/lateral impacts.
For the dynamic stimulus, where a rollout-level flag would not isolate the
edited vehicle, we instead report whether the ego footprint ever overlaps
its footprint, and its minimum clearance over the rollout.

\section{Diagnosing the Perception--Action Binding Gap}
\label{sec:diagnosis}

An unsafe plan can arise because hazard evidence is unavailable, because the
trajectory generator cannot express an adequate response, or because available
evidence has insufficient authority over the action. We distinguish these
cases explicitly.
Let $h$ denote a safety-relevant scene variable, $o$ the observation, $e$ the
ego history, $z_v(o)$ the visual representation before ego-state fusion,
$\mathrm{trace}(o)$ the model's own reasoning output, and
$z(o,e)=F(z_v(o),e,\mathrm{trace}(o))$ the representation consumed by the
trajectory generator, with $F$ the (unspecified) fusion function. Let
$q(z)$ denote its trajectory distribution. \emph{Availability} asks
whether $h$ is recoverable from $z_v(o)$; \emph{action authority} asks whether
changing $h$ while holding the remaining conditioning fixed changes the action,
\begin{equation}
\label{eq:authority}
\Delta_h(e)=
D\!\left(
q(z(o_{h_1},e)),
q(z(o_{h_0},e))
\right),
\end{equation}
where $D$ is a divergence between trajectory distributions and $h_1$, $h_0$
denote the scene with and without the hazard present, respectively, and
\emph{action-side sufficiency} asks whether the frozen generator can
produce the required safe response when an equivalent constraint is supplied
directly during generation.
We define a \emph{perception--action binding gap} as the case in which hazard evidence is available and a safe response is expressible, yet changes in that evidence exert insufficient causal influence on the generated action. In practice, $\Delta_h$ is read off simpler proxies for how much the action
moves: the sensitivity of forward reach to a swept or matched channel, that
channel's correlation with reach, and the change in footprint-intrusion
probability under matched pairs. We call authority insufficient when this
movement is small next to a channel already known to matter, and the plan
still makes contact.

\subsection{The Failure Holds Open- and Closed-Loop}
\label{sec:real_failure}
\label{sec:alpasim_closedloop}

On matched CARLA hazard scenes, contact rate, defined as footprint intrusion, is 58.1\%
\ci{56.1}{60.5} at seen sites and 65.0\% \ci{58.6}{71.1} at new sites held
out in both geometry and appearance. Note that each hazard scene is constructed with an identical twin
where just the obstacle is removed, so this gap is a causal
effect of the hazard, not a property of the site.
This result is open-loop: it scores a single predicted trajectory against a
frozen scene, so in principle the failure could vanish once the model is
allowed to observe and react as it approaches. To check this, we insert a
frozen vehicle into AlpaSim/NuRec \cite{alpasim_2025} and let the native controller replan
repeatedly as it closes the distance. The failure does not vanish; it gets
worse. Across 24 paired reconstructions, the front-collision rate rises from
about one in four rollouts to about four in five, a paired increase of
54.2 points \ci{{+}33.3}{{+}75.0}, and of the 18 scenes the unedited policy
drives cleanly, 13 collide only once the hazard is added. A control replay
of the unedited policy on these same scenes shows no comparable shift (95\%
CI includes zero), ruling out sampling noise as the explanation.

To better isolate whether this response reflects a genuine failure to avoid
a real obstacle rather than an artifact of rendering a vehicle where none
originally stood, we run a second, stricter closed-loop test that uses a
vehicle the reconstruction already contains. In the dynamic stimulus
(\Cref{sec:setup}), a logged lead vehicle either stops where the log places
it or keeps driving; only this choice differs between arms. The policy again
fails to avoid the frozen car (\Cref{tab:dynamic_closedloop}): minimum
clearance drops by a paired 23.7\,m and improves in none of the 38 scenes,
and the ego's footprint overlaps the stopped car's in most freeze rollouts
but in none of the controls.
To further isolate the braking response, we restrict our analysis to scenes
where the ego would collide with the frozen vehicle unless it brakes.
Fourteen of the 38 scenes do not meet this criterion: in eleven, the
unperturbed policy's own path already clears the frozen vehicle, and in
three, the control rollout never reaches it. In the remaining 24
``must-brake'' scenes, the policy does respond, slowing by about 2\,m/s
relative to its own control, but not enough to avoid the car: it still
contacts the stopped vehicle in 13 of the 24.

\begin{table}[t]
\caption{\textbf{The dynamic closed-loop hazard: the frozen vehicle is not
avoided.} 
Clearance is minimum separation; $\Delta$ is paired vs.\
control (95\% CI, scene-clustered bootstrap, $B{=}10^4$). The guided row is paired against the car-stops arm, i.e.\ the effect of the intervention.
The \emph{Must brake} set consists of 24 scenes where avoiding the car requires braking.}
\label{tab:dynamic_closedloop}
\centering
\scriptsize
\setlength{\tabcolsep}{2.2pt}
\begin{tabular}{@{}llccc@{}}
\toprule
 & Arm: Other & Contact $\downarrow$ & Clear. $\uparrow$ & $\Delta$ clear. \\
 & Car's Motion & (\%) & (m) & (m) \\
\midrule
\multirow{2}{*}{All ($n{=}38$)}
  & Drives on & 0.0 & 24.31 & --- \\
  & Stops & 42.1\,\ci{26.3}{57.9} & 0.09 & $-23.7$\,\ci{$-29.0$}{$-18.5$} \\
\midrule
\multirow{3}{*}{Must brake ($n{=}24$)}
  & Drives on & 0.0 & 24.31 & --- \\
  & Stops & 54.2\,\ci{33.3}{75.0} & $-0.02$ & $-23.1$\,\ci{$-28.6$}{$-17.5$} \\
  & \quad + guidance & \textbf{0.0} & \textbf{18.08} & ${+}17.3$\,\ci{{+}15.0}{{+}19.6} \\
\bottomrule
\end{tabular}
\end{table}

\subsection{Hazard Availability Does Not Imply Avoidance}
\label{sec:availability}

\begin{figure}[t]
  \centering
  \includegraphics[width=0.88\columnwidth]{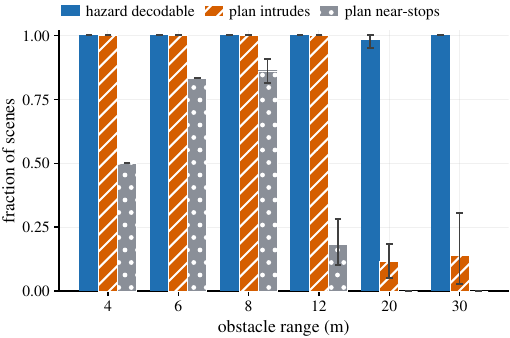}
  \caption{\textbf{Hazard presence is decodable well before the plan reacts.}
  In matched CARLA pairs, hazard presence is decodable at an area under the
  curve (AUC) of 0.998, yet every evaluated scene at 8--12\,m intrudes.
  Whiskers are 95\% CIs, clustered by site.}
  \label{fig:availability_a}
\end{figure}

To understand why the model contacts the hazard, we first test its
\emph{availability}: whether the model's visual representation encodes
the hazard at all. Using matched CARLA
despawn pairs, we probe whether hazard presence is decodable from the frozen
visual cache on the same samples where we measure the action
(\Cref{fig:availability_a}). This test is necessary because the intrusion
result alone cannot distinguish a model that fails to act on a hazard from
one that never perceives it. A linear decoder reliably separates hazard from
empty-road representations, yet every matched hazard scene at 8--12\,m still
intrudes: the evidence is present in the representation but does not change
the plan enough. Nor is the failure a simple absence of braking. Most trajectories
at 8\,m do brake, nearly to a stop, but still enter the obstacle footprint;
the response comes too late to maintain clearance.

\begin{figure}[t]
  \centering
  \includegraphics[width=0.85\columnwidth]{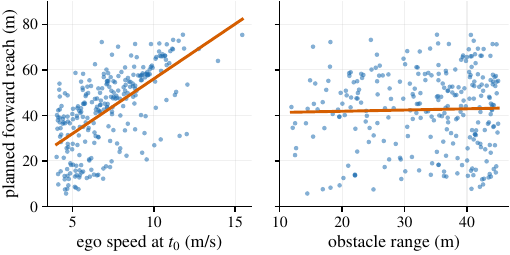}
  \caption{\textbf{The real-scene plan tracks ego speed, not obstacle range.}
  On held-out real scenes at the planner's native four-camera input, planned
  forward reach tracks logged ego speed (left) far better than privileged
  obstacle range (right).}
  \label{fig:availability_bc}
\end{figure}

The real logged scenes reproduce the same asymmetry without relying on a probe
(\Cref{fig:availability_bc}): if the plan were driven by the hazard,
obstacle range should predict forward reach at least as well
as ego speed does. Instead, range predicts reach far worse than ego speed
does, weakly enough that the range correlation is not reliably distinguishable
from zero.

\subsection{Longitudinal Authority Is Concentrated in Ego History}
\label{sec:ego_authority}

\begin{figure}[t]
  \centering
  \includegraphics[width=0.88\columnwidth]{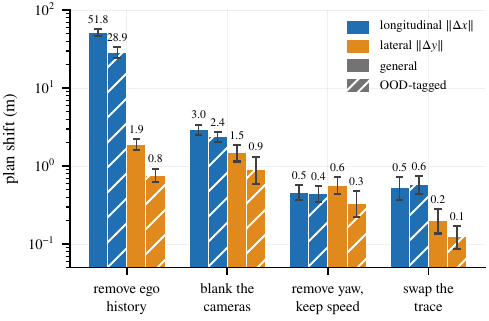}
  \caption{\textbf{Longitudinal action authority is concentrated in ego
  history.} Removing ego history changes
  longitudinal motion far more than blanking vision or the reasoning trace;
  the gap mostly disappears for lateral motion. Hatched bars repeat the
  same ablation on $n{=}120$ OOD clips. Whiskers
  are 95\% CIs bootstrapped over clips.}
  \label{fig:channels}
\end{figure}

We next test action \emph{authority} directly: what does the plan listen to
instead? We use a sample of $n{=}120$ US clips (moderate-speed driving with real curvature, not straight-line cruising).
For each clip, we run the planner with its full input,
then re-run the same state while ablating one channel at a time: blanking the
cameras, removing ego history, removing yaw while preserving speed, or
swapping the reasoning trace. We measure each channel's effect by the RMS
displacement of the resulting $K{=}6$ mean trajectory from the full-input
trajectory, separately in longitudinal and lateral motion.

The contrast is large (\Cref{fig:channels}). Removing ego history produces
by far the largest longitudinal shift, roughly an order of magnitude larger
than blanking vision, while removing yaw or swapping the reasoning trace has
still smaller effects. The imbalance is specific to longitudinal regulation:
for lateral motion, vision has substantially greater relative authority.

To check that this dominance is not an artifact of easy, uneventful driving, we
repeated the ablation on a separate $n{=}120$ sample that NVIDIA itself
flags as long-tail events: work zones, dense pedestrian traffic, and
unusual vehicle behavior. Ego history still dominates the longitudinal plan
over vision by roughly the same order of magnitude, about 12$\times$. The
absolute effect is smaller only because these scenes are driven more
slowly to begin with.

\subsection{Reasoning Has Little Longitudinal Authority}
\label{sec:cot_authority}

Ego history and vision are inputs to the frozen fusion function; the
model's own reasoning trace is a third, causally distinct channel, so we
test its action authority with two complementary interventions on different
substrates. We first swap the reasoning trace across the same $n{=}120$
U.S. clips used for the modality ablation in \Cref{fig:channels}; the
resulting longitudinal shift is the smallest of any channel tested. We then use a separate matched CARLA intervention at a fixed 8\,m range, rotating a stopped vehicle from lane-aligned to broadside while holding
site, actor, ego motion, and native four-camera input otherwise matched. This changes the fraction of traces demanding a stop from 10.4\% to 99.0\%, yet changes planned reach by only $+0.41$\,m. Because rotation also changes the observation, we treat it as
corroborating, not isolated, evidence: stated intent shifts far more than
the plan does.

\section{Restoring Perception--Action Binding at Sampling Time}
\label{sec:guidance}

Our analysis shows that the model encodes the hazard but acts on it too
late. Building on this insight, we aim to restore the missing coupling at
sampling time while keeping the model frozen. We separate enforcement from
sensing in two steps. First, we test whether an accurate safety constraint
can steer the frozen generator. Then, we replace the privileged geometry
behind that constraint with signals the model itself provides. We describe the method in detail below.

\subsection{Energy Guidance as a General Binding Intervention}
\label{subsec:target}

Let $x_k$ denote the partially integrated trajectory at flow time $t_k$, $c$
the model conditioning, and $s$ an inference-time safety signal. We
modify the flow-matching update~\cite{lipman2023flow} as
\begin{equation}
\label{eq:general_guidance}
x_{k+1}
=
x_k+\Delta t\,v_\theta(x_k,t_k;c)
-\eta_k\nabla_x E(x_k;s).
\end{equation}
Here, $E(x;s)$ converts the safety signal into a differentiable constraint
on the trajectory. The signal $s$ may originate from the visual
representation, the reasoning trace, an external oracle, or a combination;
\Cref{sec:detector_bound} later changes the source of $s$ without changing
$E$ itself.
Importantly, the energy acts only on the trajectory as it is generated, so
the intervention changes how much authority an explicitly supplied safety
signal has over the action, not what the frozen network represents.

Our diagnosis in \Cref{sec:diagnosis} localized the missing coupling to
longitudinal stopping authority, using stationary in-path obstacles as the
controlled setting. We therefore instantiate the general energy as a
longitudinal stopping energy. For planned speed $v_\tau$ and cumulative
forward reach $r_\tau$ at waypoint $\tau=1,\dots,T$, we define
\begin{equation}
\label{eq:energy}
\begin{aligned}
E_{\mathrm{stop}}(x;s)
&=
g\big\langle
w_\tau\!\left[
\sfp(v_\tau-\vtarget)^2
+\sfp(-v_\tau)^2
\right]
\big\rangle_\tau \\
&\quad+
g\lambda_{\mathrm r}
\big\langle
w_\tau\sfp(r_\tau-\dsafe)^2
\big\rangle_\tau
+
\lambda_{\mathrm j}
\big\langle(\Delta a_\tau)^2\big\rangle_\tau .
\end{aligned}
\end{equation}
The first term limits excessive acceleration and softly penalizes reversal.
The second penalizes forward motion beyond the desired stopping position and
provides the braking authority of the intervention. The final term regularizes
jerk. The fixed waypoint weights $w_\tau$ emphasize the end of the planning
horizon, while
$\eta_k=\eta_0\sin(\pi t_k)$ introduces the guidance gradually during flow
integration.

For this instantiation, the safety signal is
$s=(g,\hat d,\hat\ell)$, where $g$ is an activation gate, $\hat d$ an
estimated obstacle-center range, and $\hat\ell$ an estimated obstacle
half-length. These quantities define the stopping constraint as
\begin{equation}
\label{eq:supply}
\begin{aligned}
\dsafe
&=
\max\!\left(
\hat d-\hat\ell-L_{\mathrm{ego}}-b,\;1
\right),\\
\vtarget
&=
\max\!\left(
v_0,\sqrt{2a_{\mathrm c}[\hat d-3]_+}
\right),
\end{aligned}
\end{equation}
where $L_{\mathrm{ego}}$ accounts for the rear-axle offset and front
overhang, $b$ is a fixed safety buffer, and $a_{\mathrm c}$ a comfortable
deceleration bound. Flooring $\vtarget$ at $v_0$ keeps the speed term from
requesting deceleration on its own; braking comes entirely from the
stopping-position term. All primary experimental arms share this energy and
its hyperparameters, differing only in how $s$ is obtained, and guidance is
cheap: computing the gradient through trajectory kinematics rather than back
through the network adds 3.8\,ms, 0.2\% of the 2.30\,s base replan, at the
native $K{=}6$ budget. 

\subsection{Oracle Constraint: The Frozen Generator Can Close the Gap}
\label{subsec:oracle}
\label{subsec:threearm}
\label{subsec:oracle_closedloop}

We first test \emph{action-side sufficiency}: if the correct safety
constraint is given explicit authority during sampling, can the frozen
generator produce the missing braking response? To isolate this question,
we remove all uncertainty from the safety signal by supplying ground-truth
obstacle geometry ($g=1$, $\hat d=d^\star$, $\hat\ell=\ell^\star/2$). All
other inputs and components remain unchanged, including the visual
observations, reasoning trace, ego history, network weights, initial noise,
and sampling budget. Note that this is deliberately an oracle experiment, not a
deployable method.

\begin{figure}[t]
\centering
\includegraphics[width=0.95\columnwidth]{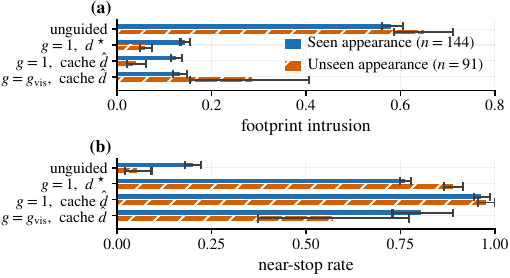}
\caption{\textbf{Exact geometry is sufficient for safe braking, and sourcing
it from the model narrows most of the gap.}
All arms share the same guidance energy and differ only in the signal
$s{=}(g,\hat d,\hat\ell)$: $d^\star$ is the privileged ground-truth range,
$\hat d$ the cache-derived estimate, and $g_{\mathrm{vis}}$ its learned gate.
\textbf{(a)} Footprint intrusion and \textbf{(b)} near-stop rate on matched
CARLA hazard sites, seen and six new sites held out in appearance. Whiskers are 95\% CIs,
site-clustered bootstraps.}
\label{fig:bottleneck}
\end{figure}

With exact geometry, footprint intrusion falls from 58.1\% to 14.5\% on seen
sites and from 65.0\% to 6.0\% on new ones, and the near-stop rate rises from
20.1\% to 76.4\% on seen sites and 5.5\% to 89.0\% on new ones
(\Cref{fig:bottleneck}a,b). The residual is a physical floor rather than
slack: on seen sites all of it sits at the 8\,m range bin, where a car or
truck leaves no stopping distance even with its exact geometry supplied, and
it is 0.0\% at 12, 20 and 30\,m. The intervention is intentionally
conservative: the experiment asks whether safe braking is \emph{expressible}
by the generator, not whether the resulting trajectory is the maneuver a
human would choose. Because the same frozen generator produces the required braking response once the constraint is given direct authority, an inability to express safe
braking cannot by itself explain the original failure.

More importantly, guidance changes the dependence that defined the binding
gap. On the same real-scene instrument, the dominant relationship with
planned reach shifts from ego speed toward obstacle range, and the
within-clip range slope goes from \NativeSlopeUnguided{}
\NativeSlopeUnguidedCI{}, statistically indistinguishable from zero, to
\NativeSlopeGuided{} \NativeSlopeGuidedCI{}, reliably positive
(\Cref{fig:coupling}). This leads to a few insights:

\begin{figure}[t]
\centering
\includegraphics[width=0.88\columnwidth]{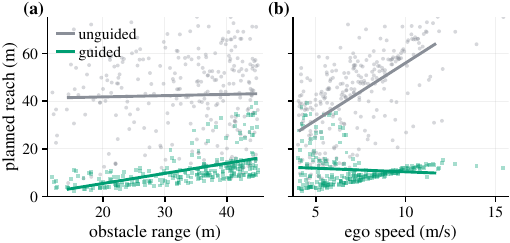}
\caption{\textbf{Guidance restores obstacle authority over longitudinal planning.}
Planned reach on \NativeSetN{} held-out real scenes vs.
\textbf{(a)}~ground-truth obstacle range and \textbf{(b)}~logged ego speed.
Without guidance (gray), reach tracks ego speed but not range
($\rho=\NativeReachSpeedRho{}$\,\NativeReachSpeedRhoCI{} vs.\
$\NativeReachRangeRho{}$\,\NativeReachRangeRhoCI{}); exact-geometry
guidance (green) reverses this, tracking range but not speed
($\rho=\NativeGuidedReachRangeRho{}$\,\NativeGuidedReachRangeRhoCI{} vs.\
$\NativeGuidedReachSpeedRho{}$\,\NativeGuidedReachSpeedRhoCI{}).}
\label{fig:coupling}
\end{figure}

\paragraph{Guidance changes the generated samples rather than merely
reranking them}
\label{sec:transport}
A cheaper alternative would be to rerank the unguided samples rather than
steer them. At the native $K{=}6$ budget, 21.5\% of real scenes have all six
unguided samples intruding, so no reranking selector could have rescued
them. Guidance clears 91\% of these outright (sample-level intrusion drops
from 98.5\% to 4.8\%): it transports trajectories during flow integration
rather than selecting one already present in the unguided set.

\paragraph{The effect persists under closed-loop}
Guidance so far was scored on logged, single-step states;
\Cref{sec:alpasim_closedloop} showed that the unguided failure actually
gets worse once the model can replan repeatedly, so we check whether the
fix survives the same test. We run it on a stricter dynamic instrument: a
lead vehicle drives its logged path, then stops at a pose it was actually
observed at, and the reconstruction renders it throughout the approach.
Where avoiding the car requires braking, the unguided model slows but does
not avoid it, striking the stopped car in 13 of 24 rollouts. Guidance
removes this entirely: minimum clearance rises from a median $-0.02$\,m to
$18.08$\,m, improving all 24 scenes; collisions fall from 13 of 24 to zero,
and every rollout comes to rest short of the car rather than driving through
it. Here the gate is deliberately fixed to $g{=}1$, so the constraint is always active. This experiment therefore isolates whether the frozen generator can enforce an accurately specified stopping constraint; deciding \emph{when} that constraint should be activated is addressed separately in the next section.

\subsection{Model-Derived Constraints: Toward Autonomous Binding}
\label{sec:detector_bound}

The oracle experiment above assumes both \emph{when} to intervene
($g{=}1$) and \emph{where} to stop (exact obstacle geometry). We now ask
how much of this safety constraint can instead be recovered from the frozen
model itself. This requires estimating both whether guidance should be
activated and where the relevant obstacle lies.

\paragraph{Safety constraints can be read from the visual representation}
We first test whether the frozen visual representation contains enough
information to replace the oracle gate and obstacle range. Using clips
disjoint from evaluation, we fit a logistic hazard-presence readout
$p_\phi(\bar z_v)$ and a ridge range readout
$\hat r_\psi(\bar z_v)$ from a pooled descriptor of the forward-view
vision tokens (mean, maximum, and standard deviation). We then replace the
oracle signal with: 
\begin{equation}
\label{eq:supply_vision}
g_{\mathrm{vis}}
=
p_\phi(\bar z_v)\,
\mathbf{1}[p_\phi(\bar z_v)>\gamma],
\qquad
\hat d=\hat r_\psi(\bar z_v),
\qquad
\hat\ell=\ell_0,
\end{equation}
where $\ell_0=2$\,m. The energy itself is unchanged.

On the designed-hazard instrument, this fully model-derived arm reduces
intrusion from 65.0\% to 28.6\% on unseen sites. Thus, signals recoverable
from the frozen representation are sufficient to recover a substantial
fraction of the oracle intervention, but a gap remains under appearance
shift. \Cref{tab:rangebottleneck} next tests where that gap comes from.

\paragraph{Autonomous guidance requires spatially resolved hazard evidence}
\label{subsec:ranging}
We next ask why model-derived guidance approaches the oracle on seen
appearances but degrades under appearance shift. We test two possible
bottlenecks: \emph{input coverage}, by adding the front-tele camera to the
front-wide input, and \emph{spatial readout}, by replacing pooled token
statistics with a small attention readout over the same frozen tokens.
All variants use the same site splits and operating threshold.

\begin{table}[t]
\caption{\textbf{Input coverage and spatial readout determine the quality
of model-derived guidance.} 
Results from the same CARLA sites as \Cref{fig:bottleneck}. 
Attention row is averaged over five seeds.
}
\label{tab:rangebottleneck}
\centering
\scriptsize
\setlength{\tabcolsep}{2.1pt}
\begin{tabular}{@{}llccccc@{}}
\toprule
 & readout & MAE (m) & recall & ceiling & intrusion & clear cost \\
 & & $\downarrow$ & $\uparrow$ & $\uparrow$ & $\downarrow$ & $\downarrow$ \\
\midrule
\multirow{3}{*}{Seen ($n{=}144$)}
  & pooled, front-wide    & \CachedRangeMAE{}     & \textbf{81\%} & \textbf{83\%} & \textbf{13.4\%} & 4.63\,m \\
  & \;+\,front-tele       & \textbf{\CachedRangeMAETele{}} & 74\% & 74\% & 16.1\% & \textbf{0.00\,m} \\
  & attention, front-wide & \CachedRangeMAEAttn{} & 76\% & 76\% & 16.3\% & \textbf{0.00\,m} \\
\midrule
\multirow{3}{*}{Unseen ($n{=}91$)}
  & pooled, front-wide    & \NewRangeMAE{}        & 55\% & 63\% & 28.6\% & \textbf{0.00\,m} \\
  & \;+\,front-tele       & \NewRangeMAETele{}    & 66\% & 68\% & 18.5\% & \textbf{0.00\,m} \\
  & attention, front-wide & \textbf{\AttnRangeMAEBare{}}   & \textbf{81\%} & \textbf{81\%} & \textbf{11.7\%} & \textbf{0.00\,m} \\
\bottomrule
\end{tabular}
\end{table}

Table~\ref{tab:rangebottleneck} isolates the source of this failure: under
appearance shift, the range readout overestimates nearby distances and
underestimates distant ones. The near-field error is the one that matters,
since it inflates the stopping distance the gate compares against and lets
real hazards slip past the intervention. This stems not from the
representation itself but from how it is read: the deployed descriptor
collapses the token grid into pooled statistics, discarding the spatial
information a distance estimate needs. Reading the same frozen tokens with a
small attention mechanism removes most of this near-field bias, and with it
most missed intrusions, without increasing false activation on hazard-free
frames.

The results identify both bottlenecks
(\Cref{tab:rangebottleneck}). Under appearance shift, adding the
front-tele camera improves intrusion from 28.6\% to 18.5\%, showing that
the front-wide view alone does not always expose the relevant in-path
obstacle strongly enough. Preserving spatial structure is even more
effective: the attention readout reduces intrusion to 11.7\% and improves
range MAE from \NewRangeMAE{} to \AttnRangeMAEBare{}\,m, without measured
reach loss on clear frames. The pooled readout loses this information
because it collapses the token grid into global statistics; under shift,
its range estimates are particularly biased in the near field, where
accurate stopping distance matters most.

The remaining errors also clarify the limit of decodability. Far-field
range error persists across readouts, consistent with limited spatial
resolution of distant tokens, while adding the tele camera and preserving
token structure largely address the near field. Thus, hazard presence being
decodable is not sufficient: autonomous guidance requires a signal that is
both spatially accurate and available in the views used by the gate.

\paragraph{Cross-domain generalization remains the bottleneck for automatic guidance}
The automatic arm generalizes across held-out CARLA sites, but not yet across
the larger CARLA-to-NuRec rendering shift encountered in closed-loop replay.
Applied unchanged to NuRec frames, the CARLA-fitted presence readout no longer separates hazard from matched-control rollouts, while range estimates are not sufficiently obstacle-specific for reliable gating. Refitting the readouts on NuRec frames improves detection and ranging, showing that the frozen representation remains usable, but exposes a deeper selectivity problem: the gate distinguishes vehicles from empty road more readily than stopped blocking vehicles from moving lead vehicles. Thus, the remaining challenge for a general automatic arm is twofold: robustness across visual domains and hazard-specific triggering within the target domain.

\paragraph{The reasoning trace provides complementary evidence for when to
intervene}
\label{subsec:moe}
We next test whether the model's reasoning provides complementary evidence
for when to activate guidance. On a separate held-out routing corpus, we parse each sampled trace into an ordered caution decision and define
\begin{equation}
\label{eq:supply_cot}
g_{\mathrm{cot}}
=
\mathbf{1}[\kappa(\mathrm{trace})\succeq\textsc{slow}],
\qquad
g_{\cup}
=
\max(g_{\mathrm{vis}},g_{\mathrm{cot}}).
\end{equation}
The range estimate $\hat d$ is unchanged from \Cref{eq:supply_vision}: the
reasoning trace only re-sources the gate.

\begin{table}[t]
\caption{\textbf{Vision and reasoning route on different evidence.}
On the routing corpus, combining vision and reasoning raises recall from
71\% to 87\%.}
\label{tab:moe}
\centering
\scriptsize
\setlength{\tabcolsep}{4pt}
\begin{tabular}{@{}lcc@{}}
\toprule
Router & Gating recall $\uparrow$ & Nominal \textsc{stop} $\downarrow$ \\
       & (\%, 95\% CI) & (\%, 95\% CI) \\
\midrule
Vision detector & 71\% \ci{62}{80} & \textbf{11\%}\,\ci{6}{16} \\
Reasoning trace & 55\% \ci{45}{65} & 15\%\,\ci{10}{21} \\
Union           & \textbf{87\%} \ci{80}{93} & 18\%\,\ci{12}{25} \\
\bottomrule
\end{tabular}
\end{table}

On the routing corpus, $g_{\cup}$ raises recall from 71\% to 87\%
(\Cref{tab:moe}), confirming that reasoning provides complementary gating
evidence, but also increases false activations. Against the stronger presence
readout used by the guidance arm, however, the trace adds only 0.9 recall
points for 16.7 points of false flags. Reasoning also provides no metric
obstacle range, so localization still depends on the visual readout.

\section{Scope Across Driving Architectures}
\label{sec:scope}

We test whether analogous hazard-response failures appear in other driving
policies beyond Alpamayo-1.5, each
exercised through its own native interface. Because these models
differ jointly in training data, sensing, action space, language use, and
ego-state conditioning, the comparisons bound the scope of the phenomenon
rather than establish a shared cause.

\paragraph{Senna}
Senna~\cite{senna2024} predicts language-level decisions rather than
trajectories, so we test whether its braking decision changes under matched
hazard and clear observations. It does not: the \textsc{decelerate+stop} rate
is 12\% in both conditions, even though the model frequently describes the
stopped vehicle in its output. Varying the supplied ego-speed token, in
contrast, changes the overall braking rate. As in Alpamayo-1.5, ego state
influences the action while the hazard itself produces little selective
response.

\paragraph{VaVAM}
VaVAM~\cite{bartoccioni2025vavim} predicts trajectories from video without
explicit ego-state conditioning. On a matched set of logged scenes scored at
a single forward camera, its footprint-intrusion rate is substantially lower
than Alpamayo-1.5's, but this does not translate into safer closed-loop
behavior: the two policies collide on the same scenes at the same rate, and
VaVAM leaves the drivable surface in most rollouts. The difference is
therefore in the form of the response, not its safety: VaVAM reacts more
laterally, Alpamayo-1.5's failure is predominantly longitudinal, and a
geometric footprint metric can count that lateral displacement as avoidance
even when the trajectory is not admissible. The same metric can be gamed
even more directly: as a stress test, we feed VaVAM a motion-free input (the
same frame repeated instead of a real clip), which drops footprint intrusion
to zero, but only because the resulting trajectories are too short to reach
the obstacle, not because the model avoids it.

Across the models tested, we find no evidence the failure is unique to
Alpamayo-1.5, though the comparisons cannot attribute a common cause.

\section{Why Standard Evaluation Can Miss a Binding Gap}
\label{sec:evaluation}

The binding gap is easy to miss because it is a
\emph{conditional-response} failure rather than simply a failure of average
prediction quality. The relevant question is not whether a policy achieves
low trajectory error or collision rate on the observed data distribution,
but whether introducing a hazard, while holding competing conditioning
fixed, causes an appropriate change in action. Standard logged evaluation
does not provide this counterfactual: hazard presence, ego motion, scene
geometry, and driver behavior vary together, so their individual influence
on the generated plan cannot in general be separated. Moreover, the critical
state studied here is itself rare in real driving data. Under our
strict filter, nuScenes contains only a handful of frames in which a
moving ego vehicle is still approaching a stationary in-path obstacle at
speed. A substantial failure of hazard-conditioned response can therefore
remain nearly invisible to aggregate open-loop or closed-loop metrics.

This motivates the designed simulation instruments used throughout the
paper. Simulation supplies the missing counterfactual: we can vary hazard
presence and obstacle range while holding scene geometry, ego motion, and
ego-history conditioning fixed, directly measuring whether hazard evidence
has causal authority over the generated action. We therefore use simulation
not to estimate the frequency of real-world collisions, but as a controlled
instrument for isolating perception--action coupling. 

\section{Limitations}
\label{sec:limitations}
Our claims have several directions for follow-up studies. Currently, every hazard we measure is synthetic: obstacles rendered in CARLA, and reconstructed vehicles re-posed or frozen in AlpaSim/NuRec. Logged driving enters this paper only as a coupling corpus, because the state we study, a moving ego that must brake for a stationary in-path obstacle, is close to absent from logged data. Future studies should validate the result in real-world, counterfactual scenarios. Second, footprint intrusion is an open-loop planning metric, not crash
probability. We therefore validate the failure in closed-loop replay, where it persists; matched A/A replays show no comparable shift. Third, the model-derived guidance arm generalizes across held-out CARLA sites, but not across the CARLA-to-NuRec rendering shift. Refitting on NuRec restores detection and range, but not the stopped-versus-moving discrimination needed for selective activation. Thus, the remaining challenge for a general
automatic arm is robust, hazard-specific triggering across domains. Fourth, decodability establishes that hazard information is recoverable from
the representation, not that it is actively used by the policy; we therefore interpret our matched CARLA probes only as evidence of availability. Finally, the cross-model study shows analogous failures beyond Alpamayo, but cannot establish a shared underlying cause.

\section{Conclusion}
\label{sec:conclusion}

We identify a \emph{perception--action binding gap} in reasoning driving
VLAs: hazard evidence is available in the visual representation and in the
reasoning trace, yet exerts little causal authority over the generated
longitudinal action.
Supplying the missing constraint during flow integration lets the same
frozen generator produce safe braking plans, cutting footprint intrusion from
58.1\% to 14.5\% and eliminating collisions entirely under closed-loop
replanning, and
reversing the diagnosed dependence from ego speed toward obstacle range. The
failure is therefore one of binding, not perception or trajectory
expressivity: the evidence is present and a safe response is available, but
the two are never connected. While the full causal diagnosis and intervention experiments are conducted
on Alpamayo-1.5, the cross-model results show that weak hazard-conditioned
action response is not unique to this architecture. Moreover, the guidance
mechanism itself is not Alpamayo-specific: it applies in principle to all VLAs with a 
flow-matching action generator.

\section*{Acknowledgment}
{\footnotesize
Generative AI (Claude, Anthropic) drafted and revised prose throughout the
main text (Sections I--IX) from author-supplied results, and wrote the
analysis and figure-generation code producing Figs.~2--6 and Tables I--III;
Fig.~1 was authored directly by the authors. Figures and tables report real
results, not AI-generated content; only their code was AI-assisted. All
AI-assisted code was executed and checked
against the underlying run data, all AI-drafted text was reviewed and
revised by the authors, and all claims were verified against those results;
the authors take responsibility for the content.}

{\scriptsize
\bibliographystyle{IEEEtran}
\bibliography{references}}

\end{document}